%% file: emnlp2020.tex
\documentclass[11pt,a4paper]{article}
\usepackage[hyperref]{emnlp2020}
\usepackage{times}
\usepackage{latexsym}

\usepackage{microtype}

\aclfinalcopy 

\usepackage{textcomp}
\usepackage{amssymb}
\usepackage{multirow}
\usepackage{adjustbox}
\usepackage{float}
\usepackage{array}
\newcolumntype{h}{>{\setbox0=\hbox\bgroup}c<{\egroup}@{}}

\title{Multi-label Few/Zero-shot Learning with Knowledge Aggregated from Multiple Label Graphs}

\author{Jueqing Lu$^1$, Lan Du$^{1}$\thanks{\,\,\,Corresponding author}, Ming Liu$^2$, and Joanna Dipnall$^3$ \\
 $^1$Faculty of Information Technology, Monash University, Australia, VIC 3800\\
 $^2$School of Information Technology, Deakin University, Australia, VIC 3217 \\
 $^3$School of Public Health and Preventive Medicine, Monash University, 
 Australia, VIC 3800 \\
 \texttt{jluu0014@student.monash.edu}\\
  \texttt{\{lan.du, jo.dipnall\}@monash.edu} \\
  \texttt{m.liu@deakin.edu.au} \\
  }

\date{}

\begin{document}
\maketitle
\begin{abstract}
Few/Zero-shot learning is a big challenge of many classifications tasks,
where a classifier is required to recognise instances of classes that
have very few or even no training samples.
It becomes more difficult in multi-label classification,
where each instance is labelled with more than one class. 
In this paper, we present a simple multi-graph aggregation model 
that fuses knowledge
from multiple label graphs encoding different semantic label relationships in order to
study how
the aggregated knowledge can benefit multi-label zero/few-shot document classification.
The model utilises three kinds of semantic information, i.e., the pre-trained 
word embeddings, label description, and pre-defined label relations.  
Experimental results derived on two large clinical datasets (i.e., MIMIC-II and MIMIC-III ) and the EU legislation dataset show that methods
equipped with the multi-graph knowledge aggregation
achieve significant performance
improvement across almost all the measures on few/zero-shot labels.
\end{abstract}

\setlength{\abovedisplayskip}{3pt}
\setlength{\belowdisplayskip}{3pt}

\section{Introduction}
\input{sec1.tex}

\section{Related Work}
\label{sec2}
\input{sec2.tex}

\section{Learning with Knowledge Aggregation}
\input{sec3.tex}

\section{Experiments}
\input{sec4.tex}

\section{Conclusion}
\label{sec:conclusion}
\input{sec5.tex}

\section*{Acknowledgments}
We thank anonymous reviewers for their valuable comments.

\bibliographystyle{acl_natbib}
\bibliography{emnlp2020}

\section*{Appendices}
\input{emnlp2020_suppl}

\end{document}

%% file: sec1.tex
Multi-label learning is a fundamental and practical problem in computer vision and natural language processing. 
Many tasks, such as automated medical coding \cite{Yan:2010:MCC,rios2018few,ocz085},
recommender systems \cite{Halder:2018}, image classification \cite{Chen_2019_CVPR,wang2019multi},
law study \cite{parikh-etal-2019-multi,chalkidis-etal-2019-large}, and stance detection \cite{ferreira-vlachos-2019-incorporating}
can be formulated as a multi-label learning problem.
Different from multi-class classification, 
an instance in multi-label learning 
is often associated with more than one class label, 
which makes the task even more challenging due to the combinatorial nature of the label space.
i.e., the number of possible label combinations is exponential with the total number of labels.

In real-world applications, 
there are often insufficient
or even unavailable
training data of ever emerging classes \cite{Vinyals:2016, xian2018zero}.
For instance,
more than half of the International Classification of Diseases (ICD) codes are not associated 
with a discharge summary in the MIMIC-III dataset \cite{johnson2016mimic,rios2018few}.
As a solution, zero-shot learning \cite{xian2018zero,Wang:2019:SZL}
aims to generalize classifiers to unseen classes by leveraging various label semantics.
Those classifiers are required to recognise instances of classes that
have never been seen in the training set,
which becomes more difficult in multi-label learning.

Moreover, the number of classes can reach hundreds of thousands.
The ICD-9-CM taxonomy contains 17K diagnosis/procedure codes\footnote{{https://www.cdc.gov/nchs/icd/icd9cm.htm}},
where the majority occurs less than 10 times in MIMIC-III;
the EU legislation corpus (EURLEX57X) \cite{chalkidis-etal-2019-large} contains about 7K labels, 
70\%of which have been assigned to less than 10 documents.
The power-law distribution of labels \cite{Liu:2017:DLE,xie2019cikm,song2019generalized}
leads to the few-shot learning challenge, 
where each label has a few training instances.

Classes come naturally with structures, 
which capture different relationships between individual classes.
For example,
codes in the ICD-9-CM taxonomy are organised 
in a rooted tree
with edges representing is-a relationships between parents and children \cite{perotte2014diagnosis}.
We can compute a code similarity graph using the code description and
a code co-occurrence graph using the annotated discharge summaries in MIMIC-II/III.
These two graphs can capture label relationships that are missing in the taxonomy.
For example, the similarity graph can reveal the relationship between ``hypertensive chronic kidney disease'' 
and ``acute kidney failure''; 
the co-occurrence graph can give us information about that ``coronary atherosclerosis of native coronary artery'' frequently co-occurs with ``coronary arteriography using two catheters''.
It has been shown that
ignoring this structured information and 
assuming all classes to be mutually exclusive are insufficient \cite{pmlr-v84-zhao18b,Gaure2017APF,kavuluru2015empirical}.

In this paper,
we present a simple but effective multi-graph knowledge aggregation
model that can transform and fuse the structural information 
from multiple label graphs while utilising three kinds of semantics:
the pre-trained word embeddings,
label description, and the label relations.
To demonstrate its efficacy,
we adapt the model as a sub-module to several existing neural architectures \cite{rios2018few,chalkidis-etal-2019-large} 
for multi-label few/zero-shot learning.
However, this model can work as a self-contained module and 
be flexibly adapted to most existing multi-label learning models \cite{xie2019cikm,li2019icd} that use GCNs to leverage the label structures.
Experiments on three real-world datasets show that neural classifiers equipped with our 
multi-graph knowledge aggregation model
can significantly improve the few/zero-shot classification performance.

%% file: sec2.tex
Leveraging structural label information via GCNs \cite{Kipf:2016tc} 
has become a promising approach of tackling the few/zero-shot problem, 
attracting increasing attention in recent years.
\citet{wang2018zero,Kampffmeyer_2019_CVPR}, and \citet{chen2017graph}
have used GCNs to learn visual classifiers for multi-class image classification.
These ideas can be generalised to multi-label learning \cite{Lee_2018_CVPR,Chen_2019_CVPR,Do2019,wang2019multi,you2019cross}.
However, none of these methods can be directly adapted to multi-label few/zero-shot text classification.
Using the label-wise attention mechanism \cite{mullenbach-etal-2018-explainable,xiao2019label},
\citet{rios2018few} introduced 
an attention-based CNN to convert each document into a feature matrix,
each row of which is a label-specific document feature vector.
The multi-label document classifiers were learned from a GCN over the label hierarchy.
While considering only the efficiency of the document encoder,
\citet{chalkidis-etal-2019-large,li2019icd,xie2019cikm} further proposed to replace the simple CNN with
BIGRU, multi-filter residual CNN and densely-connected CNN respectively.
In contrast, our work focuses on the learning of the classifiers from multiple label graphs.
Existing work on multiple graphs learning often proposed to either fuse multiple graphs
before fed into a GCN \cite{khan2019multi,wang2019multi} or consider the multi-dimensionality of graphs 
\cite{ma2019multi,Wu:2019:LDA}
for only note classification/link prediction.

%% file: sec3.tex
\paragraph{Problem Formulation} Let $\mathcal{C}_S$ and $\mathcal{C}_U$ be disjoint sets of seen and unseen labels. 
$\mathcal{C}_S$ is further divided into frequent labels $\mathcal{C}^R_S$ 
and few-shot labels $\mathcal{C}^F_S$ such that
$\mathcal{C}_S = \mathcal{C}^R_S \cup \mathcal{C}^F_S$.
Given a training set $\{(\mathbf{x}_1,\mathbf{y}_1),\dots,(\mathbf{x}_N,\mathbf{y}_N)\}$, where $\mathbf{x}_i$ indicates the $i$-th document 
and $\mathbf{y}_i \subset \mathcal{C}_S$ is the subset of labels assigned to $\mathbf{x}_i$,
the goal is to predict $\hat{\mathbf{y}}_i$ for each test document
in generalised zero-shot settings \cite{xian2018zero},
where $\hat{\mathbf{y}}_i$ is a subset of  
$\mathcal{C}_S \cup \mathcal{C}_U$.
Note that: \textit{i)} every label has a description;
\textit{ii)} the label relationships encoded in graphs can be 
computed from various resources;
\textit{iii)} documents associated with any label
from $\mathcal{C}_U$ are excluded from training.

\textbf{Document Encoder with Label-wise Attention}
According to the characteristic of different datasets,
different document encoders $\phi$ can be used to generate the document representation, i.e., $\mathbf{F}_i = \phi(\mathbf{x}_i)$.
For a corpus, like EURLEX57X,
where the average document length is in hundreds,
one can consider 
Bi-GRU/LSTM, HAN \cite{yang2016hierarchical}, BERT \cite{devlin2018bert}, etc.
For a corpus, like MIMIC-II/III,
where the discharge summaries contain multiple long and heterogeneous 
medical narratives, the CNN-based encoders have shown prominet performance,
like those discussed in Section~\ref{sec2}.

The size of $\mathbf{F}_i \in \mathbb{R}^{n\times u}$ varies, depending on the encoder.
For BERT, $n$ is the number of words and $u$ is the size of the output layer of BERT;
for CNNs, $n$ is the number of $s$-grams generated by 
CNNs with a filter size $s$ and $u$ the number of filters.

In addition, we create label embeddings $ \mathbf{v}_l$ by TF-IDF weighted average of pre-trained word embeddings
\cite{chen2017graph} according to the label description,
and use those label embeddings to compute the label-wise attention \cite{mullenbach-etal-2018-explainable, rios2018few} 
for each document $\mathbf{x}_i$
as follows:
\begin{eqnarray}
    \mathbf{a}_{i,l} &=& \textrm{softmax} (\textrm{tanh} ( \mathbf{F}_i\mathbf{W}_0 + \mathbf{b}_0)
    \mathbf{v}_l)\\
    \mathbf{z}_{i,l} &=& \mathbf{a}_{i,l}^T \mathbf{F}_i,
\end{eqnarray}
where $\mathbf{W}_0 \in \mathbb{R}^{u \times d}$, $\mathbf{b}_0 \in \mathbb{R}^ d$.
The attention is to capture how different parts of texts are relevant to different classes. 

\input{results/data_stats.tex}

\begin{figure}[!t]
    \centering
    \includegraphics[width = 0.45 \textwidth]{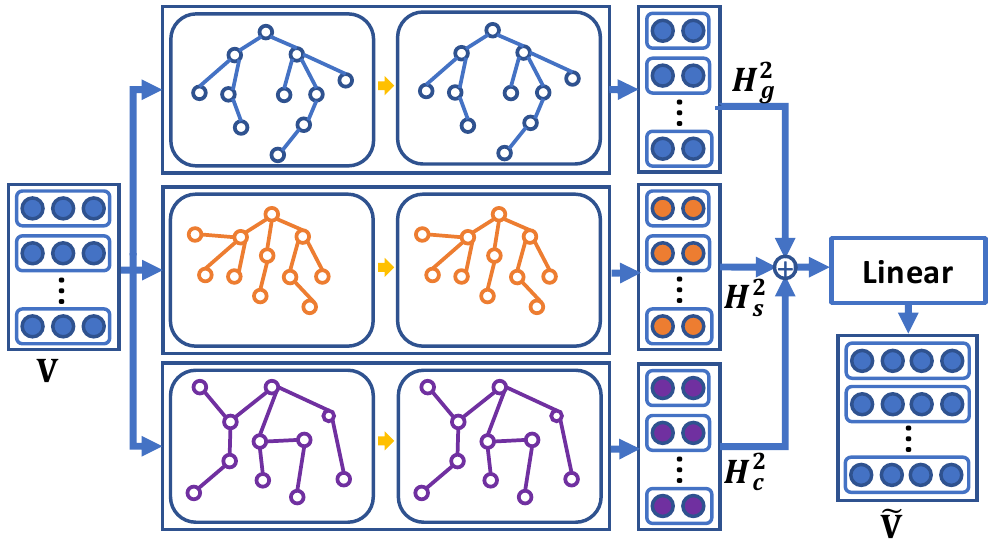}
    \caption{Multi-graph knowledge aggregation }
    \label{fig:multigcn}
    \vspace{-3mm}
\end{figure}

\textbf{Knowledge Aggregation from Multi-Graphs (KAMG) }
We consider the label hierarchy ($\mathbf{A}_g$) given by the class taxonomy,
the semantic similarity graph ($\mathbf{A}_s$) computed from their descriptions,
and the label co-occurrence graph ($\mathbf{A}_c$) 
extracted for $\mathcal{C}_S$ from the training data,
although our method can be generated to more label graphs.
Let $\mathbf{A} \in \mathbb{R}^{|\mathcal{C}_S| \times |\mathcal{C}_S|}$
be any of the three label graphs,
$\mathbf{V} \in \mathbb{R}^{L \times d}$ be the label embedding matrix,
a two-layer GCN is applied to each graph as follows:
\begin{eqnarray}
    \mathbf{H}^1 &=& \sigma(\mathbf{D}^{-1/2} \mathbf{A} \mathbf{D}^{-1/2}\mathbf{V} \mathbf{W}_1)\\
    \mathbf{H}^2 &=& \sigma(\mathbf{D}^{-1/2} \mathbf{A}\mathbf{D}^{-1/2} \mathbf{H}^1 \mathbf{W}_2)
\end{eqnarray}
where $\mathbf{D}_{i,i} = \sum_{j}A_{i,j} $ is a degree matrix of $\mathbf{A}$, $\mathbf{W}^1 \in \mathbb{R}^{d \times q}$ and $\mathbf{W}^2 \in \mathbb{R}^{q \times p}$
are two weight matrices, $\mathbf{H}^1$ and $\mathbf{H}^2$ indicate the hidden states and outputs
respectively,
$\sigma$ is the non-linear activation function,
a rectified linear unit (ReLU) in our case. 

Different from \citet{rios2018few,xie2019cikm},
we feed a two-layer GCN to each of the three graphs and 
generate three sets of label embeddings: $\mathbf{H}^2_{g}$, $\mathbf{H}^2_{s}$ and
$\mathbf{H}^2_{c}$,
which are supposed to capture different semantic relations between labels.
A linear layer is then used to fuse 
the three types of label embeddings:
\begin{equation} \label{fuse-eq}
    \tilde{\mathbf{v}}_l = f([\mathbf{h}_{g,l}^{2}, \mathbf{h}_{s,l}^{2}, \mathbf{h}_{c,l}^{2}], \mathbf{W}_3)
\end{equation}
where 
$\mathbf{W}_3 \in \mathbb{R}^{3p \times \tilde{q}}$,
and $\tilde{\mathbf{v}}_l \in \mathbb{R}^{\tilde{q}}$. 
We acknowledge that it is also worth trying the techniques used in multi-model learning
\cite{kiela2018efficient}, which is subject to future work.
Figure~\ref{fig:multigcn} visualises the multi-graph knowledge 
aggregation process.

We concatenate both
$\mathbf{v}_l$ with
$\tilde{\mathbf{v}}_l$ to form
the final text classifiers
as $\bar{\mathbf{v}}_l = [\mathbf{v}_l, \tilde{\mathbf{v}}_l],\, \bar{\mathbf{v}}_l \in \mathbb{R}^ {d+\tilde{q}}$. 
The label-wise document embeddings ($\mathbf{z}_{i,l}$) are projected
onto the same space as $\bar{\mathbf{v}}_l$ via a simple
nonlinear transformation as
\begin{equation}
\bar\mathbf{z}_{i,l} = \textrm{ReLU} (\mathbf{W}_4 \mathbf{z}_{i,l} + \mathbf{b}_4)
\end{equation}
where $\mathbf{W}_4 \in \mathbb{R}^{(d+\tilde{q}) \times u}$ and $\mathbf{b}_4 \in \mathbb{R}^{(d+\tilde{q})}$.
The prediction for each label $l$ is generated with
$\hat{y}_{i,l} = \textrm{sigmoid} (\bar{\mathbf{z}}_{i,l}^T \bar\mathbf{v}_l)$.
The model is optimised via a multi-label binary cross-entropy loss.
Although we used three label graphs (label hierarchy, similarity and co-occurrence) to demonstrate the advantage of aggregating knowledge from multi-graphs, 
the model itself is general enough to be applied to other datasets where there exist multiple label graphs.

\textbf{Zero-Shot Classification}
For zero-shot prediction, we extend $\mathbf{A} \in \mathbb{R}^{|\mathcal{C}_S| \times |\mathcal{C}_S|}$ to $\tilde{\mathbf{A}} \in \mathbb{R}^{(|\mathcal{C}_S|+|\mathcal{C}_U|) \times (|\mathcal{C}_S|+|\mathcal{C}_U|)}$, so that the new graph can encode
the relationship between unseen and seen classes. 
All labels will be optimized simultaneously during the training stage as in \cite{rios2018few}.
Note that $\mathbf{A}_c$ counts only the co-occurrence of seen classes.

%% file: results/data_stats.tex
\begin{table*}[!t] 
\footnotesize
\centering
\begin{adjustbox}{max width=0.8\textwidth}
\begin{tabular}{|c|cccccc|ccc|} \hline
& \multicolumn{6}{c|}{Docs}& \multicolumn{3}{c|}{\# Labels} \\
Dataset & \#Train & \#Dev & \#Test  & Avg \# tokens & Avg \# labels &
Voc Size & 
Frequent & \multicolumn{1}{c}{Few}  & \multicolumn{1}{c|}{Zero}   \\ \hline
MIMIC-II & 17,593 & 1,955 & 2,200 & 1,350 & 9 & 55,237
& {1,844} & \multicolumn{1}{c}{2,745}  & \multicolumn{1}{c|}{361}   \\
MIMIC-III & 47,718 & 1,631 & 3,372 & 1,931 & 15 & 104,656
& {4,204} & \multicolumn{1}{c}{4,115} & \multicolumn{1}{c|}{203} \\ 
EURLEX57K & 45,000 & 6,000 & 6,000 & 727 & 5 & 169,439
&\multicolumn{1}{c}{746} & \multicolumn{1}{c}{3,362} &\multicolumn{1}{c|}{163}  \\
\hline
\end{tabular}
\end{adjustbox}
\vspace{-3mm}
\caption{Dataset statistics}
\label{table:datalabel}
\end{table*}

%% file: sec4.tex
In this section, several experiments were conducted to
evaluate the efficacy of KAMG in classifying discharge summaries and legislative documents.
We compared our methods with several state-of-the-art multi-label classifiers in a few/zero-shot setting, and
studied how KAMG behaves by varying label graphs 
in a set of ablation experiments.

\input{results/result_table_mimic.tex}

\input{results/result_table_eu.tex}

\textbf{Datasets} 
We used two benchmark medical datasets (MIMIC II and III) and the EU legislation dataset (EURLEX57K) 
to evaluate our method in the few/zero-shot settings. 
Statistics of these datasets are shown in Table~\ref{table:datalabel}. 
Following \citet{rios2018few, chalkidis-etal-2019-large}, 
we split the datasets in such a way that 
1) zero-shot labels (i.e., unseen) do not have any instances in training;
2) few-shot labels (i.e., less frequent labels)
were defined as those whose frequencies in the training set are less than or equal to 
5 for MIMIC-II and MIMI-III 
and 50 for EURLEX57K.
The 200-dimensional word embeddings pre-trained on PubMed and MIMIC-III \cite{zhang2019biowordvec, chen2018biosentvec} 
were used for MIMIC-II/III, and 200-dimensional word embeddings pre-trained on law corpra provided by \citet{chalkidis-etal-2019-large} were used for EURLEX57k.

\textbf{Experiment settings and metrics}
For MIMIC-II/III, 
we used the NeuralClassifier \cite{liu-etal-2019-neuralclassifier} 
as a base framework
to implement our methods.
We used 200 filters with kernel size 10 to setup the CNNs by following \citet{rios2018few}
and the GCNs' hidden layer size was set to $200$.
For EURLEX57K, we leveraged \citet{chalkidis-etal-2019-large}'s code, 
and used the one-layer BiGRU with 
hidden dimension 100 as reported in their paper. 
The size of the GCNs' hidden states was set to $200$.

Moreover, the dropout rate was set to 0.2, 0.1 for MIMIC-II/III and EURLEX57K respectively and applied after the embedding layer. 
Adam optimizer
(i.e., learning rate: 0.001 for CNN and 0.0003 for BIGRU)
was used to train all the models.
All experiments were run with one NVIDIA GPU V100.

We report a variety of 
ranking metrics,
including Recall$@K$ and nDCG$@K$. 
We argue that the ranking metrics are more preferable for few/zero-shot label without introducing significant bias towards frequent labels;
they are more inline with the human annotation process, 
like the ICD coding, where clinicians often review a limited number of candidate codes.
$K$ was set to 10 for MiMIC-II/III and 5 for URLEX57K.

\textbf{Results on MIMIC-II/III}
We compared KAMG, which uses all three label graphs ($\mathbf{H}_g$, $\mathbf{H}_s$ and $\mathbf{H}_c$),
with the following baselines: 
CNN, 
RCNN (the best model in \citet{liu-etal-2019-neuralclassifier}), 
CAML, DR-CAML,
ZACNN and ZAGCNN.
Table-\ref{result_mimic} shows the performance 
of all those models. 
KAMG outperforms the other models in all 
the metrics across almost
all the settings on both datasets with a notable margin, 
due to our multi-graph knowledge aggregation model. 
Specifically, while classifying zero-shot labels,
ACNN-KAMG outperforms ZAGCNN, which uses only the label hierarchy (i.e., $\mathbf{H}_g$), by 
8\% in R$@10$ and 6.5\% in nDCG$@10$ on MIMIC-II 
and 4.1\% in R$@10$ and 10.5\% in nDCG$@10$ on MIMIC-III.
Similarly, ACNN-KAMG gains 
4\% in R$@10$ and 2.7\% in nDCG$@10$ on MIMIC-II 
and 3.7\% in R$@10$ and 6.5\% in nDCG$@10$ on MIMIC-III
over ZAGCNN on few-shot labels.

\input{results/result_table_superimposing}

\input{results/result_table_mimic_ablation}

\input{results/result_table_eu_ablation}

\textbf{Results on EURLEX57K}
We further compared AGRU-KAMG with 
with BIGRU-LAWN, ZERO-CNN-LAWN,
and ZERO-BIGRU-LAWN, which are the best performing models using label-wise attention on few/zero-shot labels in \cite{chalkidis-etal-2019-large}.
We implemented AGRU-KAMG by directly 
modifying ZERO-BIGRU-LAWN's published code.
Results in Table~\ref{result_eu} show
AGRU-KAMG performs significantly better than ZERO-BIGRU-LAWN
on zero-shot labels by gaining 
9.0\% improvement in R$@5$ and 6.9\% in nDCG$@5$,
and comparably with ZERO-BIGRU-LAWN on few-shot labels.
BIGRU-LAWN exhibits strong performance on frequent/few-shot labels, which
is inline with \citet{chalkidis-etal-2019-large}'s finding.
This could be attributed to the fine-tuning of label embeddings in the learning process.
In contrast, AGRU-KAMG has label embeddings fixed to those computed from pretrained embedding in order to leverage label description
in the zero-shot setting.

\textbf{Results on pre/post-GCN fusion}
Table~\ref{result_super} shows the performance difference between 
the following two graph fusion methods: 1) merging two label graphs into one graph, and then
feeding it into one GCN \cite{ma2019multi,wang2019multi}, and 2) our method, where two graphs were fed into two GCNs and then fused together.
The results showed that our method performs much better than the pre-GCN fusion method.

\textbf{Results on using different combinations of label graphs}
We further conducted a set of ablation experiments based on the use of different combinations of label graphs
to study how the performance of KAMG varies while using different graphs in both few and zero-shot settings.
The results in Tables~\ref{result_mimic_ablation} and \ref{result_eu_ablation} 
show that i) KAMG performs better with multiple graphs than with a single graph overall, which demonstrates
it is beneficial to aggregate information from multiple graphs;
ii) graphs contribute differently to the classification performance,
the ICD taxonomy plays an important role while being used in conjunction with the other graphs, and the three graphs work complementary to each other
on EURLEX57K.

%% file: results/result_table_mimic.tex
\begin{table*}[!t]
\centering
\footnotesize
\begin{adjustbox}{width=\textwidth}
\begin{tabular}{|c|l|chc|chc|chc|chc|}
\hline
\multicolumn{2}{|c|}{}
    & \multicolumn{3}{c|}{Frequent} 
    & \multicolumn{3}{c|}{Few} 
    & \multicolumn{3}{c|}{Zero} 
    & \multicolumn{3}{c|}{Overall} 
\\ \cline{3-14}
    \multicolumn{2}{|c|}{}
    & R@10 & RP@10 & nDCG@10 
    & R@10 & RP@10 & nDCG@10 
    & R@10 & RP@10 & nDCG@10
    & R@10 & RP@10 & nDCG@10
\\ \hline 
    \parbox[t]{2mm}{\multirow{7}{*}{\rotatebox[origin=c]{90}{MIMIC-II}}}
&CNN \cite{kim2014convolutional}
    & 0.346 & 0.408 & 0.465 
    & 0.032 & 0.032 & 0.018 
    & - & - & - 
    & 0.335 & 0.400 & 0.460
\\
&RCNN \cite{lai2015recurrent}
    & 0.386 & 0.451 & 0.505 
    & 0.081 & 0.081 & 0.047 
    & - & - & - 
    & 0.373 & 0.442 & 0.498
\\\cline{2-14}
&CAML \cite{mullenbach-etal-2018-explainable}
    &0.386 & 0.455 & 0.508 
    & 0.078 & 0.078 & 0.043 
    & 0.021 & 0.021 & 0.012 
    & 0.371 & 0.445 & 0.501
\\
&DR-CAML \cite{mullenbach-etal-2018-explainable}
    & 0.383 & 0.451 & 0.502 
    & 0.075 & 0.075 & 0.044 
    & 0.028 & 0.028 & 0.016 
    & 0.368 & 0.441 & 0.495
\\
&ZACNN \cite{rios2018few}
    & 0.445 & 0.521 & 0.562 
    & 0.180 & 0.180 & 0.114 
    & 0.362 & 0.362 & 0.225 
    & 0.424 & 0.506 & 0.551
\\
&ZAGCNN \cite{rios2018few}
    & \textbf{0.471} & 0.551 & \textbf{0.591}
    & 0.219 & 0.219 & 0.139 
    & 0.382 & 0.382 & 0.231 
    & \textbf{0.452} & 0.538 & \textbf{0.583}
\\

& ACNN-KAMG  
    & \textbf{0.471} & 0.551 & \textbf{0.591} 
    & \textbf{0.259} & \textbf{0.259} & \textbf{0.166}
    & \textbf{0.462} & \textbf{0.4617} & \textbf{0.296} 
    & 0.451 & 0.538 & 0.582
\\\hline \hline
\parbox[t]{2mm}{\multirow{7}{*}{\rotatebox[origin=c]{90}{MIMIC-III}}}
&CNN \cite{kim2014convolutional}
    & 0.366 & 0.577 & 0.632
    & 0.074 & 0.074 & 0.044
    & - & - & - 
    & 0.361 & 0.575 & 0.631
\\
&RCNN \cite{lai2015recurrent}
    & 0.376 & 0.591 & 0.648 
    & 0.118 & 0.118 & 0.070 
    & - & - & -  
    & 0.370 & 0.588 & 0.646
\\\cline{2-14}
&CAML \cite{mullenbach-etal-2018-explainable}
    & 0.422 & 0.662 & 0.711 
    & 0.104 & 0.104 & 0.073 
    & 0.067 & 0.067 & 0.029 
    & 0.415 & 0.659 & 0.709
\\
&DR-CAML \cite{mullenbach-etal-2018-explainable}
    & 0.416 & 0.652 & 0.699
    & 0.105 & 0.105 & 0.064
    & 0.038 & 0.038 & 0.018 
    & 0.409 & 0.649 & 0.697
\\
&ZACNN \cite{rios2018few}
    & 0.405 & 0.636 & 0.684
    & 0.207 & 0.207 & 0.104
    & 0.457 & 0.457 & 0.222 
    & 0.372 & 0.595 & 0.654
\\
&ZAGCNN \cite{rios2018few}
    & 0.427 & 0.668 & 0.713 
    & 0.258 & 0.258 & 0.130 
    & 0.512 & 0.512 & 0.253 
    & 0.394 & 0.629 & 0.685
\\
& ACNN-KAMG  
    & \textbf{0.434} & \textbf{0.679} & \textbf{0.724} 
    & \textbf{0.295} & \textbf{0.295} & \textbf{0.195} 
    & \textbf{0.553} & 0.553 & \textbf{0.358}
    & \textbf{0.427} & 0.676 & \textbf{0.722}
\\ \hline
\end{tabular}
\end{adjustbox}
\vspace{-3mm}
\caption{Multi-label classification results on MIMIC-II and MIMIC-III. Bold figures indicate the best results for each score.}
\label{result_mimic}
\vspace{-3mm}
\end{table*}

%% file: results/result_table_eu.tex
\begin{table*}[!t]
\centering
\footnotesize
\begin{adjustbox}{max width=\textwidth}
\begin{tabular}{|l|chc|chc|chc|chc|}
\hline
\multicolumn{1}{|c|}{}
    & \multicolumn{3}{c|}{Frequent} 
    & \multicolumn{3}{c|}{Few} 
    & \multicolumn{3}{c|}{Zero} 
    & \multicolumn{3}{c|}{Overall} 
\\ \cline{2-13}
    \multicolumn{1}{|c|}{}
    & R@5 & RP@5 & nDCG@5
    & R@5 & RP@5 & nDCG@5
    & R@5 & RP@5 & nDCG@5
    & R@5 & RP@5 & nDCG@5
\\ \hline 

 BIGRU-LWAN \cite{chalkidis-etal-2019-large}
    & \textit{0.755} & - & \textit{0.819}
    & \textit{0.661} & - & \textit{0.618}
    & 0.029 & - & 0.019
    & \textit{0.692} & - & \textit{0.796}
 \\\cline{1-13}
  ZERO-CNN-LWAN \cite{chalkidis-etal-2019-large}
    & 0.683 & - & 0.745
    & 0.494 & - & 0.454
    & 0.321 & - & 0.264
    & 0.617 & - & 0.717
\\
  ZERO-BIGRU-LWAN \cite{chalkidis-etal-2019-large}
    & 0.716 & - & 0.780
    & 0.560 & - & 0.510
    & 0.438 & - & 0.345
    & 0.648 & - & 0.752
\\
  AGRU-KAMG 
    & \textbf{0.731} & - & \textbf{0.795}
    & \textbf{0.563} & - & \textbf{0.518}
    & \textbf{0.528} & - & \textbf{0.414}
    & \textbf{0.661} & - & \textbf{0.766}
\\\hline
\end{tabular}
\end{adjustbox}
\vspace{-3mm}
\caption{Multi-label classification results on EURLEX57K. Bold figures indicate the best results for each score 
among the three models designed specifically for zero-shot learning. Italics indicate the best results overall.}
\label{result_eu}
\vspace{-3mm}
\end{table*}

%% file: results/result_table_superimposing.tex
\begin{table*}[!ht]
\centering
\footnotesize
\begin{adjustbox}{width=\textwidth}
\begin{tabular}{|c|l|chc|chc|chc|chc|}
\hline
\multicolumn{2}{|c|}{}
    & \multicolumn{3}{c|}{Frequent} 
    & \multicolumn{3}{c|}{Few} 
    & \multicolumn{3}{c|}{Zero} 
    & \multicolumn{3}{c|}{Overall} 
\\ \cline{3-14}
    \multicolumn{2}{|c|}{}
    & R@10 & RP@10 & nDCG@10 
    & R@10 & RP@10 & nDCG@10 
    & R@10 & RP@10 & nDCG@10
    & R@10 & RP@10 & nDCG@10
\\ \hline 
    \parbox[t]{2mm}{\multirow{2}{*}{\rotatebox[origin=c]{90}{MIMIC-II}}}
& ACNN-KAMG ($\mathbf{H}_g,\mathbf{H}_s$)      
    & \textbf{0.477} & \textbf{0.558} & \textbf{0.597}
    & \textbf{0.274} & 0.274 & \textbf{0.180}
    & \textbf{0.451} & 0.451 & \textbf{0.301} 
    & \textbf{0.457} & \textbf{0.544} & \textbf{0.588}
\\
& ACNN-KAMG ($\mathbf{H}_{g + s}$)     
    & 0.470 &  & 0.587
    & 0.235 &  &  0.151
    & 0.418 &  &  0.273
    & 0.450 &  & 0.578
\\\cline{2-14}
& ACNN-KAMG ($\mathbf{H}_g, \mathbf{H}_c$)  
    & \textbf{0.476} & 0.556 & \textbf{0.596}
    & \textbf{0.277} & \textbf{0.277} & \textbf{0.177}
    & \textbf{0.454} & 0.454 & \textbf{0.282}
    & \textbf{0.456} & 0.542 & \textbf{0.586}
\\
& ACNN-KAMG ($\mathbf{H}_{g+c}$)     
    & 0.467 &  & 0.586
    & 0.236 &  &  0.152
    & 0.417 &  &  0.267
    & 0.448 &  & 0.577
\\\hline \hline
\parbox[t]{2mm}{\multirow{4}{*}{\rotatebox[origin=c]{90}{MIMIC-III}}}
& ACNN-KAMG ($\mathbf{H}_g, \mathbf{H}_s$)     
    & \textbf{0.435} & \textbf{0.680} & \textbf{0.725} 
    & \textbf{0.293} & 0.293 & \textbf{0.193} 
    & 0.530 & 0.530 & \textbf{0.346}
    & \textbf{0.428} & \textbf{0.677} & \textbf{0.723}
\\
& ACNN-KAMG ($\mathbf{H}_{g + s}$)     
    & 0.426 &  & 0.712
    & 0.256 &  & 0.130
    & \textbf{0.540} &  &  0.273
    & 0.393 &  & 0.684
\\\cline{2-14}
& ACNN-KAMG ($\mathbf{H}_g,\mathbf{H}_c$)  
    & \textbf{0.432} & \textbf{0.676} & \textbf{0.721} 
    & \textbf{0.284} & 0.284 & \textbf{0.192} 
    & \textbf{0.560} & \textbf{0.560} & \textbf{0.370} 
    & \textbf{0.425} & 0.673 & \textbf{0.720}
\\
& ACNN-KAMG ($\mathbf{H}_{g + c}$)     
    & 0.422 &  & 0.707
    & 0.245 &  &  0.123
    & 0.521 &  &  0.265
    & 0.392 &  & 0.680
\\ \hline
\end{tabular}
\end{adjustbox}
\vspace{-3mm}
\caption{The comparison of the knowledge fusion before and after GCN on MIMIC-II and MIMIC-III. Bold figures indicate the best results for each score}
\label{result_super}
\vspace{-3mm}
\end{table*}

%% file: results/result_table_mimic_ablation.tex
\begin{table*}[!t]
\centering
\footnotesize
\begin{adjustbox}{width=\textwidth}
\begin{tabular}{|l|chc|chc|chc|chc|}
\hline
\multicolumn{1}{|c|}{}
    & \multicolumn{6}{c|}{MIMIC-II} 
    & \multicolumn{6}{c|}{MIMIC-III} 
\\ \cline{2-13}
\multicolumn{1}{|c|}{}
    & \multicolumn{3}{c|}{Few} 
    & \multicolumn{3}{c|}{Zero} 
    & \multicolumn{3}{c|}{Few} 
    & \multicolumn{3}{c|}{Zero} 
\\ \cline{2-13}
    \multicolumn{1}{|c|}{}
    & R@10 & RP@10 & nDCG@10 
    & R@10 & RP@10 & nDCG@10
    & R@10 & RP@10 & nDCG@10 
    & R@10 & RP@10 & nDCG@10
\\ \hline 
ACNN-KAMG ($\mathbf{H}_g$)
    & 0.219 & 0.219 & 0.139 
    & 0.382 & 0.382 & 0.231 
    & 0.258 & 0.258 & 0.130 
    & 0.512 & 0.512 & 0.253 
\\
ACNN-KAMG ($\mathbf{H}_s$)
    & 0.245 &  & 0.157
    & 0.437 &  & 0.272 
    & 0.258 & - & 0.130
    & 0.524 & - & 0.258
\\
ACNN-KAMG ($\mathbf{H}_c$)
    & 0.248 &  & 0.157 
    & 0.424 &  & 0.267 
    & 0.252 &  & 0.130
    & 0.518 &  & 0.256 
\\
ACNN-KAMG ($\mathbf{H}_c$, $\mathbf{H}_s$)
    & 0.257 &  & 0.161
    & 0.439 &  & 0.286 
    & 0.252 &  & 0.138 
    & 0.533 &  & 0.267 
\\
ACNN-KAMG ($\mathbf{H}_g,\mathbf{H}_s$)      
    & 0.274 & 0.274 & \textbf{0.180}
    & 0.451 & 0.451 & \textbf{0.301}
    & 0.293 & 0.293 & 0.193 
    & 0.530 & 0.530 & 0.346 
\\
ACNN-KAMG ($\mathbf{H}_g, \mathbf{H}_c$)  
    & \textbf{0.277} & \textbf{0.277} & 0.177 
    & 0.454 & 0.454 & 0.282 
    & 0.284 & 0.284 & 0.192 
    & \textbf{0.560} & \textbf{0.56}0 & \textbf{0.370} 
\\
ACNN-KAMG ($\mathbf{H}_g, \mathbf{H}_s, \mathbf{H}_c$)  
    & 0.259 & 0.259 & 0.166
    & \textbf{0.462} & \textbf{0.462} & 0.296 
    & \textbf{0.295} & \textbf{0.295} & \textbf{0.195}
    & 0.553 & 0.553 & 0.358 
\\\hline \hline
\end{tabular}
\end{adjustbox}
\vspace{-3mm}
\caption{Ablation study on MIMIC-II and MIMIC-III. We ran ACNN-KAMG with different combinations of the three graphs in the few/zero-shot setting. Bold figures indicate the best results for each score.}
\label{result_mimic_ablation}
\vspace{-3mm}
\end{table*}

%% file: results/result_table_eu_ablation.tex
\begin{table*}[!t]
\centering
\footnotesize
\begin{adjustbox}{max width=\textwidth}
\begin{tabular}{|l|hhhchc|chc|hhh}
\cline{1-10}
\multicolumn{1}{|c|}{}
    & \multicolumn{3}{h}{Frequent} 
    & \multicolumn{3}{c|}{Few} 
    & \multicolumn{3}{c|}{Zero} 
    & \multicolumn{3}{h}{Overall} 
\\ \cline{2-10}
    \multicolumn{1}{|c|}{}
    & R@5 & RP@5 & nDCG@5
    & R@5 & RP@5 & nDCG@5
    & R@5 & RP@5 & nDCG@5
    & R@5 & RP@5 & nDCG@5
\\ \cline{1-10}
 AGRU-KAMG ($\mathbf{H}_g$) 
    & 0.696 & - & 0.760
    & 0.474 & - & 0.431
    & 0.472 & - & 0.363
    & 0.625 & - & 0.729
 \\
 AGRU-KAMG ($\mathbf{H}_s$) 
    & 0.707 & - & 0.771
    & 0.508 & - & 0.464
    & 0.484 & - & 0.382
    &  & - &
\\
AGRU-KAMG ($\mathbf{H}_c$) 
    & 0.708 & - & 0.771
    & 0.503 & - & 0.459
    & 0.491 & - & 0.381
    &  & - & 
 \\
AGRU-KAMG ($\mathbf{H}_c, \mathbf{H}_s$)      
    & 0.726 & - & 0.790
    & 0.554 & - & 0.509
    & 0.499 & - & 0.397
    &  & - & 
\\
AGRU-KAMG ($\mathbf{H}_g, \mathbf{H}_s$)      
    & 0.727 & - & 0.791
    & 0.550 & - & 0.504
    & 0.480 & - & 0.381
    & 0.656 & - & 0.761
\\
AGRU-KAMG ($\mathbf{H}_g, \mathbf{H}_c$)  
    & 0.727 & - & 0.792
    & 0.554 & - & 0.507
    & 0.517 & - & \textbf{0.422}
    & 0.657 & - & 0.763
\\
AGRU-KAMG ($\mathbf{H}_g, \mathbf{H}_s, \mathbf{H}_c$)  
    & 0.731 & - & 0.795
    & \textbf{0.563} & - & \textbf{0.518}
    & \textbf{0.528} & - & 0.414
    & 0.661 & - & 0.766
\\\cline{1-10}
\end{tabular}
\end{adjustbox}
\vspace{-3mm}
\caption{Ablation study on EURLEX57K. We ran AGRU-KAMG with different combinations of the three graphs in the few/zero-shot setting. Bold figures indicate the best results for each score.}
\label{result_eu_ablation}
\vspace{-3mm}
\end{table*}

%% file: sec5.tex
We have proposed a multi-graph 
aggregation method that can effectively fuse
knowledge from multiple label graphs.
Experiments on MIMIC-II/III and EURLEX57K have shown that the classifiers
derived from the multi-graph aggregation have achieved substantial performance
improvements particularly on few/zero-shot labels.
As future work, we will further study our method's ability of extreme multi-label learning \cite{Bhatia16} and different document encoders.

%% file: emnlp2020_suppl.tex
Tables~\ref{full_rak}, \ref{full_pak}, \ref{full_rpak} and \ref{full_ndcg} 
present a full set of experiments results
computed with different metrics, including, Recall$@K$, Precision$@K$, Recall-Precision$@K$,
nDCG$@K$.
All the experiments were run on one NVIDIA GPU V100.

\input{results/full_results_rak.tex}
\input{results/full_results_ndcgak.tex}
\input{results/full_results_pak.tex}
\input{results/full_results_rpak.tex}

%% file: results/full_results_rak.tex
\begin{table*}[!t]
\centering
\small
\begin{adjustbox}{max width=\textwidth}
\begin{tabular}{|c|l|ccc|ccc|ccc|ccc|}
\hline
\multicolumn{2}{|c|}{}
    & \multicolumn{3}{c|}{Frequent} 
    & \multicolumn{3}{c|}{Few} 
    & \multicolumn{3}{c|}{Zero} 
    & \multicolumn{3}{c|}{Overall}
\\ \cline{3-14}
 
    \multicolumn{2}{|c|}{}
    & R@1 & R@5 & R@10
    & R@1 & R@5 & R@10
    & R@1 & R@5 & R@10
    & R@1 & R@5 & R@10
\\ \hline 
    \parbox[t]{2mm}{\multirow{16}{*}{\rotatebox[origin=c]{90}{MIMIC-II}}}
&CNN 
    & 0.080 & 0.253 & 0.346 & 0.005 & 0.021 & 0.032 & - & - & - & 0.077 & 0.245 & 0.335
\\
&RCNN 
    & 0.086 & 0.277 & 0.386 & 0.015 & 0.048 & 0.081 & - & - & - & 0.083 & 0.267 & 0.372 \\
&CAML
    & 0.082 & 0.278 & 0.386 & 0.014 & 0.043 & 0.078 & 0.004 & 0.014 & 0.021 & 0.079 & 0.267 & 0.371 \\
&DR-CAML
    & 0.080 & 0.276 & 0.383 & 0.016 & 0.046 & 0.075 & 0.005 & 0.019 & 0.028 & 0.077 & 0.265 & 0.368 \\

&ZACNN
    & 0.086 & 0.308 & 0.445 & 0.050 & 0.126 & 0.180 & 0.101 & 0.262 & 0.362 & 0.082 & 0.294 & 0.424 \\
&ZAGCNN
   & 0.089 & 0.323 & 0.471 & 0.060 & 0.161 & 0.219 & 0.102 & 0.267 & 0.382 & 0.085 & 0.309 & 0.452 \\ \cline{2-14}

& ACNN-KAMG($\mathbf{H}_g$)
    & 0.089 & 0.319 & 0.467 & 0.066 & 0.172 & 0.235 & 0.141 & 0.302 & 0.402 & 0.085 & 0.305 & 0.448 \\
& ACNN-KAMG($\mathbf{H}_s$)
    & 0.088 & 0.322 & 0.469 & 0.069 & 0.178 & 0.245 & 0.126 & 0.315 & 0.437 & 0.084 & 0.308 & 0.449 \\
& ACNN-KAMG($\mathbf{H}_c$)
    & 0.088 & 0.323 & 0.474 & 0.068 & 0.178 & 0.247 & 0.127 & 0.305 & 0.424 & 0.084 & 0.309 & 0.454 \\
& ACNN-KAMG ($\mathbf{H}_{g+s}$)
    & 0.088 & 0.320 & 0.470 & 0.067 & 0.171 & 0.235 & 0.140 & 0.323 & 0.418 & 0.084 & 0.307 & 0.450 \\
& ACNN-KAMG ($\mathbf{H}_{g+c}$)
    & 0.088 & 0.319 & 0.467 & 0.068 & 0.177 & 0.236 & 0.136 & 0.308 & 0.416 & 0.084 & 0.306 & 0.448 \\

& ACNN-KAMG ($ \mathbf{H}_g, \mathbf{H}_s$)
    & 0.090 & 0.325 & 0.477 & 0.083 & 0.203 & 0.274 & 0.163 & 0.345 & 0.451 & 0.086 & 0.311 & 0.457 \\
& ACNN-KAMG ($ \mathbf{H}_g, \mathbf{H}_c$)
    & 0.091 & 0.325 & 0.476 & 0.077 & 0.200 & 0.277 & 0.130 & 0.323 & 0.454 & 0.086 & 0.311 & 0.456 \\
& ACNN-KAMG ($ \mathbf{H}_c, \mathbf{H}_s$)
    & 0.091 & 0.324 & 0.475 & 0.067 & 0.177 & 0.248 & 0.137 & 0.343 & 0.447 & 0.086 & 0.310 & 0.454 \\
& ACNN-KAMG ($ \mathbf{H}_g, \mathbf{H}_s, \mathbf{H}_c$)
    & 0.089 & 0.322 & 0.471 & 0.072 & 0.188 & 0.259 & 0.145 & 0.342 & 0.462 & 0.085 & 0.309 & 0.451 \\ \hline \hline
    
    \parbox[t]{2mm}{\multirow{16}{*}{\rotatebox[origin=c]{90}{MIMIC-III}}}
&CNN 
    & 0.061 & 0.240 & 0.366 & 0.017 & 0.051 & 0.074 & - & - & - & 0.060 & 0.236 & 0.361 \\
&RCNN 
    & 0.063 & 0.247 & 0.376 & 0.027 & 0.080 & 0.118 & - & - & - & 0.062 & 0.243 & 0.370 \\
&CAML
    & 0.066 & 0.267 & 0.422 & 0.038 & 0.084 & 0.104 & 0.002 & 0.036 & 0.067 & 0.065 & 0.262 & 0.415\\
&DR-CAML
    & 0.065 & 0.263 & 0.416 & 0.026 & 0.073 & 0.105 & 0.003 & 0.016 & 0.038 & 0.063 & 0.258 & 0.409\\
&ZACNN
    & 0.064 & 0.256 & 0.405 & 0.008 & 0.140 & 0.207 & 0.007 & 0.309 & 0.457 & 0.063 & 0.241 & 0.372\\
&ZAGCNN
    & 0.065 & 0.266 & 0.427 & 0.006 & 0.181 & 0.258 & 0.007 & 0.367 & 0.512 & 0.064 & 0.252 & 0.394\\
 \cline{2-14}

& ACNN-KAMG($\mathbf{H}_s$)
    & 0.065 & 0.262 & 0.420 & 0.004 & 0.184 & 0.258 & 0.007 & 0.376 & 0.524 & 0.063 & 0.247 & 0.385 \\
& ACNN-KAMG($\mathbf{H}_c$)
    & 0.065 & 0.262 & 0.419 & 0.007 & 0.171 & 0.252 & 0.007 & 0.374 & 0.518 & 0.063 & 0.245 & 0.382 \\
& ACNN-KAMG ($\mathbf{H}_{g+s}$)
    & 0.065 & 0.265 & 0.426 & 0.009 & 0.181 & 0.256 & 0.007 & 0.401 & 0.540 & 0.064 & 0.251 & 0.393 \\
& ACNN-KAMG ($\mathbf{H}_{g+c}$)
    & 0.065 & 0.263 & 0.422 & 0.008 & 0.166 & 0.245 & 0.007 & 0.397 & 0.521 & 0.064 & 0.250 & 0.392 \\

& ACNN-KAMG ($ \mathbf{H}_g, \mathbf{H}_s$)
    & 0.066 & 0.271 & 0.435 & 0.101 & 0.224 & 0.293 & 0.172 & 0.412 & 0.530 & 0.065 & 0.266 & 0.428 \\
& ACNN-KAMG ($ \mathbf{H}_g, \mathbf{H}_c$)
    & 0.066 & 0.270 & 0.432 & 0.103 & 0.216 & 0.284 & 0.194 & 0.449 & 0.560 & 0.065 & 0.265 & 0.425 \\
& ACNN-KAMG ($ \mathbf{H}_c, \mathbf{H}_s$)
    & 0.066 & 0.268 & 0.423 & 0.052 & 0.192 & 0.280 & 0.021 & 0.386 & 0.566 & 0.065 & 0.263 & 0.414 \\
& ACNN-KAMG ($ \mathbf{H}_g, \mathbf{H}_s, \mathbf{H}_c$)
    & 0.066 & 0.271 & 0.434 & 0.096 & 0.231 & 0.295 & 0.180 & 0.417 & 0.553 & 0.065 & 0.266 & 0.427

\\\hline \hline
\parbox[t]{2mm}{\multirow{7}{*}{\rotatebox[origin=c]{90}{EU}}}
&AGRU-KAMG ($ \mathbf{H}_g$)
    & 0.229 & 0.696 & 0.836 
    & 0.282	& 0.474&	0.550&	0.226&	0.472&	0.551&	0.194&	0.625&	0.762
\\
&AGRU-KAMG ($ \mathbf{H}_c$)
    & 0.232&	0.708&	0.847&	0.303&	0.503&	0.585&	0.254&	0.491&	0.574&	0.196&	0.636&	0.775
\\

&AGRU-KAMG ($\mathbf{H}_s$)
    &0.231&	0.707&	0.847&	0.305&	0.508&	0.586&	0.258	&0.484	&0.593&	0.197&	0.636&	0.776 
\\
&AGRU-KAMG ($ \mathbf{H}_c, \mathbf{H}_s$)
    & 0.237	&0.726&	0.868	&0.316&	0.554&	0.630&	0.267	&0.499&	0.606&	0.201&	0.656&	0.796
\\

&AGRU-KAMG ($ \mathbf{H}_g, \mathbf{H}_s$)
    & 0.238	&0.727	&0.864	&0.333&	0.550	&0.631&	0.257	&0.480	&0.569&	0.201&	0.656&	0.795
\\
&AGRU-KAMG ($ \mathbf{H}_g, \mathbf{H}_c$)
    & 0.238 & 	0.727 & 	0.868 & 	0.335 & 	0.554 & 	0.628 & 	0.298 & 	0.517 & 	0.641 & 	0.201 & 	0.657	 & 0.799
\\
&AGRU-KAMG ($ \mathbf{H}_g, \mathbf{H}_s, \mathbf{H}_c$)
    & 0.238&	0.731&	0.869&	0.342&	0.563	&0.643&	0.268&	0.528	&0.635&	0.201&	0.661	& 0.801
\\\hline
\end{tabular}
\end{adjustbox}
\caption{Recall@k results on MIMIC-II, MIMIC-III and EURLEX57K (EU) datasets}
\label{full_rak}
\end{table*}

%% file: results/full_results_ndcgak.tex
\begin{table*}[!t]
\centering
\begin{adjustbox}{max width=\textwidth}
\begin{tabular}{|c|l|ccc|ccc|ccc|ccc|}
\hline
\multicolumn{2}{|c|}{}
    & \multicolumn{3}{c|}{Frequent} 
    & \multicolumn{3}{c|}{Few} 
    & \multicolumn{3}{c|}{Zero} 
    & \multicolumn{3}{c|}{Overall}
\\ \cline{3-14}
    \multicolumn{2}{|c|}{}
    & nDCG@1 & nDCG@5 & nDCG@10
& nDCG@1 & nDCG@5 & nDCG@10
& nDCG@1 & nDCG@5 & nDCG@10
& nDCG@1 & nDCG@5 & nDCG@10
\\ \hline 
    \parbox[t]{2mm}{\multirow{16}{*}{\rotatebox[origin=c]{90}{MIMIC-II}}}
&CNN 
    & 0.712 & 0.538 & 0.465 & 0.007 & 0.014 & 0.018 & - & - & - & 0.711 & 0.536 & 0.460 \\

&RCNN 
    & 0.739 & 0.574 & 0.505 & 0.022 & 0.035 & 0.047 & - & - & - & 0.738 & 0.572 & 0.498 \\
&CAML
    & 0.727 & 0.578 & 0.508 & 0.018 & 0.031 & 0.043 & 0.004 & 0.009 & 0.012 & 0.726 & 0.576 & 0.501 \\
&DR-CAML
    & 0.713 & 0.571 & 0.502 & 0.023 & 0.034 & 0.044 & 0.005 & 0.013 & 0.016 & 0.712 & 0.569 & 0.495 \\

&ZACNN
    & 0.752 & 0.619 & 0.562 & 0.066 & 0.095 & 0.114 & 0.114 & 0.191 & 0.225 & 0.750 & 0.615 & 0.551 \\
&ZAGCNN
    & 0.778 & 0.648 & 0.591 & 0.077 & 0.119 & 0.139 & 0.118 & 0.193 & 0.231 & 0.777 & 0.645 & 0.583 \\
\cline{2-14}
& ACNN-KAMG($\mathbf{H}_g$)
    & 0.777 & 0.641 & 0.586 & 0.084 & 0.128 & 0.151 & 0.160 & 0.231 & 0.264 & 0.776 & 0.638 & 0.578 \\
& ACNN-KAMG($\mathbf{H}_s$)
    & 0.772 & 0.644 & 0.588 & 0.090 & 0.133 & 0.157 & 0.143 & 0.231 & 0.272 & 0.770 & 0.641 & 0.578 \\

& ACNN-KAMG($\mathbf{H}_c$)
    & 0.772 & 0.645 & 0.591 & 0.088 & 0.133 & 0.157 & 0.141 & 0.227 & 0.267 & 0.770 & 0.642 & 0.581 \\
& ACNN-KAMG ($\mathbf{H}_{g+s}$)
    & 0.770 & 0.642 & 0.587 & 0.086 & 0.129 & 0.151 & 0.155 & 0.241 & 0.273 & 0.769 & 0.639 & 0.578 \\
& ACNN-KAMG ($\mathbf{H}_{g+c}$)
    & 0.769 & 0.641 & 0.585 & 0.087 & 0.132 & 0.152 & 0.153 & 0.231 & 0.267 & 0.768 & 0.638 & 0.577 \\
& ACNN-KAMG ($ \mathbf{H}_g, \mathbf{H}_s$)
    & 0.784 & 0.652 & 0.597 & 0.109 & 0.155 & 0.180 & 0.186 & 0.266 & 0.301 & 0.783 & 0.649 & 0.588 \\
& ACNN-KAMG ($ \mathbf{H}_g, \mathbf{H}_c$)
    & 0.785 & 0.650 & 0.596 & 0.100 & 0.150 & 0.177 & 0.146 & 0.238 & 0.282 & 0.784 & 0.647 & 0.586 \\
& ACNN-KAMG ($ \mathbf{H}_c, \mathbf{H}_s$)
    & 0.785 & 0.649 & 0.595 & 0.085 & 0.132 & 0.157 & 0.159 & 0.251 & 0.286 & 0.783 & 0.646 & 0.585 \\
& ACNN-KAMG ($ \mathbf{H}_g, \mathbf{H}_s, \mathbf{H}_c$)
    & 0.780 & 0.647 & 0.591 & 0.092 & 0.141 & 0.166 & 0.165 & 0.256 & 0.296 & 0.778 & 0.644 & 0.581 \\
\hline \hline
    \parbox[t]{2mm}{\multirow{16}{*}{\rotatebox[origin=c]{90}{MIMIC-III}}}
&CNN 
    & 0.826 & 0.720 & 0.632 & 0.020 & 0.036 & 0.044 & - & - & - & 0.826 & 0.719 & 0.631
\\
&RCNN 
    & 0.845 & 0.739 & 0.648 & 0.034 & 0.057 & 0.070 & - & - & - & 0.845 & 0.738 & 0.646
\\
&CAML
    & 0.884 & 0.788 & 0.711 & 0.045 & 0.066 & 0.073 & 0.007 & 0.019 & 0.029 & 0.884 & 0.787 & 0.709
\\
&DR-CAML
    & 0.859 & 0.775 & 0.699 & 0.032 & 0.053 & 0.064 & 0.005 & 0.010 & 0.018 & 0.859 & 0.775 & 0.697

\\
&ZACNN
    & 0.858 & 0.762 & 0.684 & 0.010 & 0.081 & 0.104 & 0.007 & 0.173 & 0.222 & 0.858 & 0.748 & 0.654
\\
&ZAGCNN
    & 0.875 & 0.786 & 0.713 & 0.007 & 0.103 & 0.130 & 0.007 & 0.205 & 0.253 & 0.875 & 0.774 & 0.685
\\
 \cline{2-14}
& ACNN-KAMG($\mathbf{H}_s$)
    & 0.872 & 0.778 & 0.703 & 0.005 & 0.105 & 0.130 & 0.007 & 0.210 & 0.258 & 0.872 & 0.765 & 0.673
\\
& ACNN-KAMG($\mathbf{H}_c$)
    & 0.873 & 0.778 & 0.703 & 0.008 & 0.098 & 0.126 & 0.007 & 0.209 & 0.256 & 0.873 & 0.761 & 0.668

\\
& ACNN-KAMG ($\mathbf{H}_{g+s}$)
    & 0.874 & 0.784 & 0.712 & 0.009 & 0.105 & 0.130 & 0.007 & 0.227 & 0.272 & 0.873 & 0.773 & 0.683
\\
& ACNN-KAMG ($\mathbf{H}_{g+c}$)
    & 0.873 & 0.780 & 0.707 & 0.009 & 0.096 & 0.123 & 0.007 & 0.223 & 0.265 & 0.873 & 0.769 & 0.680
\\
& ACNN-KAMG ($ \mathbf{H}_g, \mathbf{H}_s$)
    & 0.885 & 0.797 & 0.725 & 0.118 & 0.169 & 0.193 & 0.190 & 0.307 & 0.346 & 0.885 & 0.797 & 0.723
\\
& ACNN-KAMG ($ \mathbf{H}_g, \mathbf{H}_c$)
    & 0.883 & 0.795 & 0.721 & 0.120 & 0.169 & 0.192 & 0.215 & 0.333 & 0.370 & 0.882 & 0.794 & 0.719
\\
& ACNN-KAMG ($ \mathbf{H}_c, \mathbf{H}_s$)
    & 0.884 & 0.792 & 0.713 & 0.059 & 0.128 & 0.159 & 0.028 & 0.221 & 0.280 & 0.884 & 0.791 & 0.709
\\
& ACNN-KAMG ($ \mathbf{H}_g, \mathbf{H}_s, \mathbf{H}_c$)
    & 0.882 & 0.797 & 0.724 & 0.109 & 0.172 & 0.195 & 0.203 & 0.313 & 0.358 & 0.882 & 0.796 & 0.722

\\\hline \hline
    \parbox[t]{2mm}{\multirow{7}{*}{\rotatebox[origin=c]{90}{EU}}}
    &AGRU-KAMG ($ \mathbf{H}_g$)
    & 0.857	& 0.760& 0.805& 	0.415& 	0.431& 	0.460& 	0.247& 	0.363& 	0.388& 	0.862& 	0.729& 	0.760
\\
&AGRU-KAMG ($ \mathbf{H}_c$)
    & 0.865& 0.771& 	0.816& 	0.444& 	0.459& 	0.490& 	0.272& 	0.381& 	0.410& 	0.871& 	0.740& 	0.772
\\

&AGRU-KAMG ($\mathbf{H}_s$)
    &0.866& 0.771& 	0.815& 	0.447& 	0.464& 	0.493& 	0.276& 	0.382& 	0.420& 	0.873& 	0.740& 	0.773
\\
&AGRU-KAMG ($ \mathbf{H}_c, \mathbf{H}_s$)
    & 0.881	& 0.790& 	0.834& 	0.496& 	0.509& 	0.538& 	0.285& 	0.397& 	0.432& 	0.889& 	0.761& 	0.793
\\
&AGRU-KAMG ($ \mathbf{H}_g, \mathbf{H}_s$)
    & 0.882& 	0.791& 	0.834& 	0.489& 	0.504& 	0.534& 	0.267& 	0.381& 	0.409& 	0.888& 	0.761& 	0.793
\\
&AGRU-KAMG ($ \mathbf{H}_g, \mathbf{H}_c$)
    & 0.884& 	0.792& 	0.837& 	0.491& 	0.507& 	0.535& 	0.323& 0.422& 	0.462& 	0.891& 	0.763& 	0.796
\\
&AGRU-KAMG ($\mathbf{H}_g,\mathbf{H}_s,\mathbf{H}_c$)
    & 0.883& 	0.795& 	0.839& 	0.504& 	0.518& 	0.548& 	0.290& 	0.414& 	0.447& 	0.891& 0.766& 0.798
\\\hline
\end{tabular}
\end{adjustbox}
\vspace{-3mm}
\caption{nDCG@k results on MIMIC-II, MIMIC-III and EURLEX57K (EU) datasets}
\label{full_ndcg}
\vspace{-5mm}
\end{table*}

%% file: results/full_results_pak.tex
\begin{table*}[!t]
\centering
\small
\begin{adjustbox}{max width=\textwidth}
\begin{tabular}{|c|l|ccc|ccc|ccc|ccc|}
\hline
\multicolumn{2}{|c|}{}
    & \multicolumn{3}{c|}{Frequent} 
    & \multicolumn{3}{c|}{Few} 
    & \multicolumn{3}{c|}{Zero} 
    & \multicolumn{3}{c|}{Overall}
\\ \cline{3-14}
 
    \multicolumn{2}{|c|}{}
    & P@1 & P@5 & P@10
    & P@1 & P@5 & P@10
    & P@1 & P@5 & P@10
    & P@1 & P@5 & P@10
\\ \hline 
    \parbox[t]{2mm}{\multirow{16}{*}{\rotatebox[origin=c]{90}{MIMIC-II}}}
&CNN 
    & 0.712 & 0.478 & 0.337 & 0.007 & 0.006 & 0.004 & - & - & - & 0.711 & 0.477 & 0.337 \\
&RCNN 
    & 0.739 & 0.513 & 0.369 & 0.022 & 0.014 & 0.012 & - & - & - & 0.738 & 0.512 & 0.369 \\
&CAML
    & 0.727 & 0.522 & 0.378 & 0.018 & 0.012 & 0.011 & 0.004 & 0.003 & 0.003 & 0.726 & 0.521 & 0.377 \\

&DR-CAML
    & 0.713 & 0.517 & 0.374 & 0.023 & 0.014 & 0.011 & 0.005 & 0.004 & 0.003 & 0.712 & 0.517 & 0.373 \\

&ZACNN
    & 0.752 & 0.568 & 0.429 & 0.066 & 0.034 & 0.025 & 0.114 & 0.062 & 0.043 & 0.750 & 0.566 & 0.426 \\
&ZAGCNN
    & 0.778 & 0.596 & 0.454 & 0.077 & 0.043 & 0.030 & 0.118 & 0.063 & 0.046 & 0.777 & 0.595 & 0.453 \\
 \cline{2-14}
& ACNN-KAMG($\mathbf{H}_g$)
    & 0.777 & 0.588 & 0.450 & 0.084 & 0.046 & 0.032 & 0.160 & 0.070 & 0.048 & 0.776 & 0.587 & 0.450 \\

& ACNN-KAMG($\mathbf{H}_s$)
    & 0.772 & 0.593 & 0.451 & 0.090 & 0.047 & 0.034 & 0.143 & 0.074 & 0.052 & 0.770 & 0.592 & 0.450 \\
& ACNN-KAMG($\mathbf{H}_c$)
    & 0.772 & 0.594 & 0.454 & 0.088 & 0.048 & 0.034 & 0.141 & 0.071 & 0.050 & 0.770 & 0.593 & 0.453 \\
& ACNN-KAMG ($\mathbf{H}_{g+s}$)
    & 0.770 & 0.590 & 0.451 & 0.086 & 0.044 & 0.031 & 0.155 & 0.075 & 0.050 & 0.769 & 0.589 & 0.450 \\
& ACNN-KAMG ($\mathbf{H}_{g+c}$)
    & 0.769 & 0.590 & 0.450 & 0.087 & 0.046 & 0.032 & 0.153 & 0.071 & 0.049 & 0.768 & 0.589 & 0.449 \\
& ACNN-KAMG ($ \mathbf{H}_g, \mathbf{H}_s$)
    & 0.784 & 0.599 & 0.458 & 0.109 & 0.054 & 0.037 & 0.186 & 0.080 & 0.054 & 0.783 & 0.598 & 0.457 \\
& ACNN-KAMG ($ \mathbf{H}_g, \mathbf{H}_c$)
    & 0.785 & 0.597 & 0.456 & 0.100 & 0.053 & 0.038 & 0.146 & 0.077 & 0.054 & 0.784 & 0.596 & 0.456 \\
& ACNN-KAMG ($ \mathbf{H}_c, \mathbf{H}_s$)
    & 0.785 & 0.595 & 0.455 & 0.085 & 0.047 & 0.033 & 0.159 & 0.081 & 0.055 & 0.783 & 0.594 & 0.454 \\

& ACNN-KAMG ($ \mathbf{H}_g, \mathbf{H}_s, \mathbf{H}_c$)
    & 0.780 & 0.595 & 0.453 & 0.092 & 0.051 & 0.035 & 0.165 & 0.081 & 0.056 & 0.778 & 0.594 & 0.452 \\
\hline \hline
    \parbox[t]{2mm}{\multirow{16}{*}{\rotatebox[origin=c]{90}{MIMIC-III}}}
&CNN 
    & 0.826 & 0.684 & 0.548 & 0.020 & 0.012 & 0.009 & - & - & - & 0.826 & 0.684 & 0.548 \\
&RCNN 
    & 0.845 & 0.702 & 0.560 & 0.034 & 0.021 & 0.016 & - & - & - & 0.845 & 0.701 & 0.560 \\
&CAML
    & 0.884 & 0.754 & 0.628 & 0.045 & 0.022 & 0.014 & 0.007 & 0.009 & 0.008 & 0.884 & 0.754 & 0.628 \\
&DR-CAML
    & 0.859 & 0.744 & 0.618 & 0.032 & 0.018 & 0.014 & 0.005 & 0.004 & 0.005 & 0.859 & 0.744 & 0.618 \\
&ZACNN
    & 0.858 & 0.728 & 0.603 & 0.010 & 0.035 & 0.026 & 0.007 & 0.069 & 0.052 & 0.858 & 0.710 & 0.567 \\
&ZAGCNN
    & 0.875 & 0.755 & 0.633 & 0.007 & 0.044 & 0.032 & 0.007 & 0.085 & 0.059 & 0.875 & 0.739 & 0.599 \\
 \cline{2-14}
& ACNN-KAMG($\mathbf{H}_s$)
    & 0.872 & 0.746 & 0.623 & 0.005 & 0.045 & 0.032 & 0.007 & 0.086 & 0.060 & 0.872 & 0.728 & 0.586 \\
& ACNN-KAMG($\mathbf{H}_c$)
    & 0.873 & 0.745 & 0.621 & 0.008 & 0.040 & 0.031 & 0.007 & 0.084 & 0.059 & 0.873 & 0.723 & 0.581 \\
& ACNN-KAMG ($\mathbf{H}_{g+s}$)
    & 0.874 & 0.753 & 0.632 & 0.009 & 0.044 & 0.032 & 0.007 & 0.092 & 0.061 & 0.873 & 0.738 & 0.599 \\
& ACNN-KAMG ($\mathbf{H}_{g+c}$)
    & 0.873 & 0.747 & 0.626 & 0.009 & 0.040 & 0.030 & 0.007 & 0.089 & 0.058 & 0.873 & 0.733 & 0.596 \\
& ACNN-KAMG ($ \mathbf{H}_g, \mathbf{H}_s$)
     & 0.885 & 0.766 & 0.645 & 0.118 & 0.054 & 0.036 & 0.190 & 0.094 & 0.060 & 0.885 & 0.766 & 0.645 \\
& ACNN-KAMG ($ \mathbf{H}_g, \mathbf{H}_c$)
    & 0.883 & 0.763 & 0.641 & 0.120 & 0.053 & 0.036 & 0.215 & 0.103 & 0.064 & 0.882 & 0.763 & 0.641 \\
& ACNN-KAMG ($ \mathbf{H}_g, \mathbf{H}_c$)
    & 0.884 & 0.759 & 0.629 & 0.059 & 0.046 & 0.034 & 0.028 & 0.088 & 0.064 & 0.884 & 0.758 & 0.627\\
& ACNN-KAMG ($ \mathbf{H}_g, \mathbf{H}_s, \mathbf{H}_c$)
    & 0.882 & 0.766 & 0.643 & 0.109 & 0.055 & 0.037 & 0.203 & 0.095 & 0.063 & 0.882 & 0.766 & 0.643 \\
\cline{2-14}
\hline \hline
\parbox[t]{2mm}{\multirow{7}{*}{\rotatebox[origin=c]{90}{EU}}}
&AGRU-KAMG ($ \mathbf{H}_g$)
    & 0.857& 	0.581& 	0.361& 	0.415& 	0.158& 	0.094& 	0.247& 	0.103& 	0.060& 	0.862& 	0.596& 	0.375
\\
&AGRU-KAMG ($ \mathbf{H}_c$)
    &0.865&	0.590&	0.366&	0.438&	0.167&	0.099&	0.272&	0.105&	0.062&	0.871&	0.607&	0.382
\\

&AGRU-KAMG ($\mathbf{H}_s$)
    & 0.866	&0.588&	0.366&	0.447&	0.168&	0.099&	0.276&	0.105&	0.064&	0.873&	0.625&	0.382
\\
&AGRU-KAMG ($ \mathbf{H}_c, \mathbf{H}_s$)
    & 0.881&	0.606&	0.373&	0.496&	0.184&	0.107&	0.285&	0.108&	0.065&	0.889&	0.626&	0.393
\\
&AGRU-KAMG ($ \mathbf{H}_g, \mathbf{H}_s$)
    & 0.882	&0.606&	0.373&	0.489&	0.183&	0.107&	0.276&	0.105&	0.062&	0.888&	0.626&	0.392
\\
&AGRU-KAMG ($ \mathbf{H}_g, \mathbf{H}_c$)
    & 0.884& 0.607& 0.375& 	0.491& 	0.184& 	0.107& 	0.323& 0.112& 	0.069& 	0.891& 	0.627& 	0.394
\\
&AGRU-KAMG ($ \mathbf{H}_g, \mathbf{H}_s, \mathbf{H}_c$)
    & 0.883	 & 0.610 & 0.376 & 	0.504 & 	0.188 & 	0.110 & 	0.290 & 0.115 & 	0.068 & 	0.891 & 	0.630 & 	0.396
\\\hline
\end{tabular}
\end{adjustbox}
\caption{Precision@k results on MIMIC-II, MIMIC-III and EURLEX57K (EU) datasets}
\label{full_pak}
\end{table*}

%% file: results/full_results_rpak.tex
\begin{table*}[!t]
\centering
\small
\begin{adjustbox}{max width=\textwidth}
\begin{tabular}{|c|l|ccc|ccc|ccc|ccc|}
\hline
\multicolumn{2}{|c|}{}
    & \multicolumn{3}{c|}{Frequent} 
    & \multicolumn{3}{c|}{Few} 
    & \multicolumn{3}{c|}{Zero} 
    & \multicolumn{3}{c|}{Overall}
\\ \cline{3-14}
 
    \multicolumn{2}{|c|}{}
    & RP@1 & RP@5 & RP@10
    & RP@1 & RP@5 & RP@10
    & RP@1 & RP@5 & RP@10
    & RP@1 & RP@5 & RP@10
\\ \hline 
    \parbox[t]{2mm}{\multirow{16}{*}{\rotatebox[origin=c]{90}{MIMIC-II}}}
&CNN 
    & 0.712 & 0.478 & 0.337 & 0.007 & 0.006 & 0.004 & - & - & - & 0.711 & 0.477 & 0.337 \\

&RCNN 
    & 0.739 & 0.513 & 0.369 & 0.022 & 0.014 & 0.012 & - & - & - & 0.738 & 0.512 & 0.369 \\
&CAML
    & 0.727 & 0.522 & 0.378 & 0.018 & 0.012 & 0.011 & 0.004 & 0.003 & 0.003 & 0.726 & 0.521 & 0.377 \\

&DR-CAML
    & 0.713 & 0.517 & 0.374 & 0.023 & 0.014 & 0.011 & 0.005 & 0.004 & 0.003 & 0.712 & 0.517 & 0.373 \\

&ZACNN
    & 0.752 & 0.568 & 0.429 & 0.066 & 0.034 & 0.025 & 0.114 & 0.062 & 0.043 & 0.750 & 0.566 & 0.426 \\
&ZAGCNN
    & 0.778 & 0.596 & 0.454 & 0.077 & 0.043 & 0.030 & 0.118 & 0.063 & 0.046 & 0.777 & 0.595 & 0.453 \\
\cline{2-14}
& ACNN-KAMG($\mathbf{H}_g$)
  & 0.777 & 0.603 & 0.547 & 0.084 & 0.173 & 0.235 & 0.160 & 0.303 & 0.402 & 0.776 & 0.600 & 0.534 \\

& ACNN-KAMG($\mathbf{H}_s$)
    & 0.772 & 0.609 & 0.549 & 0.090 & 0.178 & 0.245 & 0.143 & 0.315 & 0.437 & 0.770 & 0.604 & 0.535 \\
& ACNN-KAMG($\mathbf{H}_c$)
    & 0.772 & 0.610 & 0.554 & 0.088 & 0.178 & 0.247 & 0.141 & 0.305 & 0.424 & 0.770 & 0.606 & 0.539 \\
& ACNN-KAMG ($\mathbf{H}_{g+s}$)
   & 0.770 & 0.606 & 0.549 & 0.086 & 0.171 & 0.235 & 0.155 & 0.324 & 0.418 & 0.769 & 0.602 & 0.535 \\
& ACNN-KAMG ($\mathbf{H}_{g+c}$)
    & 0.769 & 0.605 & 0.547 & 0.087 & 0.178 & 0.236 & 0.153 & 0.308 & 0.416 & 0.768 & 0.601 & 0.534 \\

& ACNN-KAMG ($ \mathbf{H}_g, \mathbf{H}_s$)
    & 0.784 & 0.615 & 0.558 & 0.109 & 0.203 & 0.274 & 0.186 & 0.346 & 0.451 & 0.783 & 0.611 & 0.544 \\
& ACNN-KAMG ($ \mathbf{H}_g, \mathbf{H}_c$)
    & 0.785 & 0.613 & 0.556 & 0.100 & 0.200 & 0.277 & 0.146 & 0.324 & 0.454 & 0.784 & 0.609 & 0.542 \\
& ACNN-KAMG ($ \mathbf{H}_c, \mathbf{H}_s$)
    & 0.785 & 0.611 & 0.555 & 0.085 & 0.177 & 0.248 & 0.159 & 0.344 & 0.447 & 0.783 & 0.607 & 0.540 \\
& ACNN-KAMG ($ \mathbf{H}_g, \mathbf{H}_s, \mathbf{H}_c$)
    & 0.780 & 0.610 & 0.551 & 0.092 & 0.188 & 0.259 & 0.165 & 0.344 & 0.462 & 0.778 & 0.606 & 0.538\\
\hline \hline
    \parbox[t]{2mm}{\multirow{16}{*}{\rotatebox[origin=c]{90}{MIMIC-III}}}
&CNN 
    & 0.826 & 0.688 & 0.577 & 0.020 & 0.051 & 0.074 & - & - & - & 0.826 & 0.687 & 0.575 \\
&RCNN 
    & 0.845 & 0.706 & 0.591 & 0.034 & 0.080 & 0.118 & - & - & - & 0.845 & 0.705 & 0.588 \\
&CAML
    & 0.884 & 0.759 & 0.662 & 0.045 & 0.084 & 0.104 & 0.007 & 0.036 & 0.067 & 0.884 & 0.758 & 0.659 \\

&DR-CAML
    & 0.859 & 0.749 & 0.652 & 0.032 & 0.073 & 0.105 & 0.005 & 0.016 & 0.038 & 0.859 & 0.749 & 0.649 \\

&ZACNN
    & 0.858 & 0.733 & 0.635 & 0.010 & 0.140 & 0.207 & 0.007 & 0.309 & 0.457 & 0.858 & 0.714 & 0.595 \\
&ZAGCNN
    & 0.875 & 0.759 & 0.668 & 0.007 & 0.181 & 0.258 & 0.007 & 0.367 & 0.512 & 0.875 & 0.743 & 0.629 \\
 \cline{2-14}

& ACNN-KAMG($\mathbf{H}_s$)
    & 0.872 & 0.750 & 0.657 & 0.005 & 0.184 & 0.258 & 0.007 & 0.376 & 0.524 & 0.872 & 0.732 & 0.615 \\
& ACNN-KAMG($\mathbf{H}_c$)
    & 0.873 & 0.750 & 0.656 & 0.008 & 0.171 & 0.252 & 0.007 & 0.374 & 0.518 & 0.873 & 0.727 & 0.610 \\
& ACNN-KAMG ($\mathbf{H}_{g+s}$)
    & 0.874 & 0.757 & 0.667 & 0.009 & 0.181 & 0.256 & 0.007 & 0.401 & 0.540 & 0.873 & 0.741 & 0.628 \\
& ACNN-KAMG ($\mathbf{H}_{g+c}$)
    & 0.873 & 0.752 & 0.661 & 0.009 & 0.167 & 0.245 & 0.007 & 0.397 & 0.521 & 0.873 & 0.737 & 0.625 \\

& ACNN-KAMG ($ \mathbf{H}_g, \mathbf{H}_s$)
    & 0.885 & 0.771 & 0.680 & 0.118 & 0.224 & 0.293 & 0.190 & 0.412 & 0.530 & 0.885 & 0.770 & 0.677 \\
& ACNN-KAMG ($ \mathbf{H}_g, \mathbf{H}_c$)
    & 0.883 & 0.768 & 0.676 & 0.120 & 0.217 & 0.284 & 0.215 & 0.449 & 0.560 & 0.882 & 0.768 & 0.673 \\
& ACNN-KAMG ($ \mathbf{H}_c, \mathbf{H}_s$)
    & 0.884 & 0.763 & 0.663 & 0.059 & 0.192 & 0.280 & 0.028 & 0.386 & 0.566 & 0.884 & 0.762 & 0.658 \\
& ACNN-KAMG ($ \mathbf{H}_g, \mathbf{H}_s, \mathbf{H}_c$)
    & 0.882 & 0.770 & 0.679 & 0.109 & 0.231 & 0.295 & 0.203 & 0.417 & 0.553 & 0.882 & 0.770 & 0.675 \\
\cline{2-14}

\hline \hline
    \parbox[t]{2mm}{\multirow{7}{*}{\rotatebox[origin=c]{90}{EU}}}
    &AGRU-KAMG ($ \mathbf{H}_g$)
    & 0.857&	0.743&	0.836&	0.415&	0.475&	0.550&	0.247&	0.472&	0.551&	0.862&	0.692&	0.762
\\
&AGRU-KAMG ($ \mathbf{H}_c$)
    & 0.865 &	0.755 &	0.847 &	0.444 &	0.504 &	0.585 &	0.272 &	0.488 &	0.574 &	0.871 &	0.705 &	0.775
\\

&AGRU-KAMG ($\mathbf{H}_s$)
    & 0.866	 &0.755	 &0.847	 &0.447	 &0.509 &	0.586 &	0.276 &	0.477 &	0.595 &	0.873 &	0.705 &	0.776
\\
&AGRU-KAMG ($ \mathbf{H}_c, \mathbf{H}_s$)
    & 0.881	 &0.774	 &0.865	 &0.496	 &0.555 &	0.630 &	0.285 &	0.499 &	0.606 &	0.889 &	0.726 &	0.796
\\
&AGRU-KAMG ($ \mathbf{H}_g, \mathbf{H}_s$)
    & 0.882	& 0.778& 	0.858& 	0.489& 	0.551& 	0.631& 	0.276& 	0.480& 	0.569& 	0.888& 	0.734& 	0.795
\\
&AGRU-KAMG ($ \mathbf{H}_g, \mathbf{H}_c$)
    &  0.884&	0.776&	0.868&	0.491&	0.555&	0.628&	0.323&	0.517&	0.641&	0.891&	0.728&	0.799
\\
&AGRU-KAMG ($ \mathbf{H}_g, \mathbf{H}_s, \mathbf{H}_c$)
    & 0.883	&0.780	&0.870&	0.504&	0.564&	0.643&	0.290&	0.528&	0.635&	0.891&	0.732&	0.802
\\\hline
\end{tabular}
\end{adjustbox}
\vspace{-3mm}
\caption{R-Precision@k results on MIMIC-II, MIMIC-III and EURLEX57K (EU) datasets}
\label{full_rpak}
\vspace{-5mm}
\end{table*}